# Deep Learning for Epidemiologists: An Introduction to Neural Networks

Stylianos Serghiou, Kathryn Rough

# Footnotes page

## Abbreviations

AUPRC = area under the precision-recall curve; AUROC = area under the receiver operating characteristic; CNN = convolutional neural network; DL = deep learning; EHR = electronic health record; FNN = feed-forward neural network; GRU = gated recurrent unit; LSTM = long short-term memory; ML = machine learning; ReLU = rectified linear unit; RGB = red green blue; RNN = recurrent neural network; SELU = Scaled Exponential Linear Unit

## Corresponding author

Stylianos Serghiou, MD, PhD Google Health, 1600 Amphitheatre Parkway, Mountain View, CA 94043, serghiou@google.com

# Abstract

Deep learning methods are increasingly being applied to problems in medicine and healthcare. However, few epidemiologists have received formal training in these methods. To bridge this gap, this article introduces to the fundamentals of deep learning from an epidemiological perspective. Specifically, this article reviews core concepts in machine learning (overfitting, regularization, hyperparameters), explains several fundamental deep learning architectures (convolutional neural networks, recurrent neural networks), and summarizes training, evaluation, and deployment of models. We aim to enable  the reader to engage with and critically evaluate medical applications of  deep learning, facilitating a dialogue between computer scientists and epidemiologists that will improve the safety and efficacy of applications of this technology.

In 1998, researchers used longitudinal data on seven carefully curated predictors from 5,345 individuals to build the Framingham Risk Score for 10-year risk of coronary heart disease (1). In the subsequent decades, it became one of the most well-known and frequently utilized medical risk prediction models. Investigators prospectively collected data from a 12-year cohort study and fit a Cox model to predictors identified using decades of domain-specific knowledge. Though not typically regarded as such, the Framingham Risk Score and many other popular scores (2–4) are early applications of machine learning in medicine (5).

Deep learning is a subset of machine learning methods, recent advancements in which have led to breakthroughs in tasks that are not easily handled by more traditional methods, including image recognition (6,7), language translation (8), text-to-speech generation (9), and text synthesis (10,11). While some deep learning techniques were originally proposed in the 1980s, the increased availability of large datasets (12) and of better computing resources (13), have led to dramatic performance improvements in recent years.

Deep learning techniques are increasingly being applied to tasks in the health and medical domains (14), though few have reached the stage of clinical implementation (15). Studies demonstrate that models can use optical coherence tomography images to classify need for clinical referral on par with retinal specialists (16), use chest X-rays to diagnose pneumonia on par with radiologists (17), and use electronic health records (EHR) to predict acute kidney injury two days in advance (18), among many other applications (19,20). These results have led to excitement, despite few prospective evaluations and several methodological concerns (21–24).

With the abundance of research in this domain, it is important that epidemiologists and other health researchers are able to critically engage with and contribute to research using deep learning. This review offers an accessible introduction to the basics of deep learning from an epidemiologic perspective. It covers fundamental principles of machine learning, an explanation of common deep learning architectures, and summarizes training, evaluation, and deployment of models.

## MACHINE LEARNING FUNDAMENTALS

In the mid-80s, researchers at Massachusetts General Hospital developed a system that ranked possible diagnoses based on up to 4,700 signs or symptoms. One of the first automated decision support systems of its kind, DXplain (25,26) was based on human-curated rules and information. For example, an expert specified chest pain and shortness of breath could be signs of myocardial infarction, pulmonary embolism, aortic dissection, or other conditions.

DXplain is an 'expert system', a form of artificial intelligence (AI) outside the domain of machine learning (Figure 1). 'Artificial intelligence' (27,28) was a phrase coined in the 1950s to encompass a broad collection of machine behavior and abilities traditionally attributed to intelligent beings, including image recognition, text summarization, and commonsense reasoning. The term 'AI' does not place constraints on the methods used to achieve these goals. Early work in the field of AI, including DXplain, focused on creating decision systems that followed hard-coded logical rules. In contrast, machine learning is a data-driven approach to AI that relies on "the ability to learn without being explicitly programmed" (a quote often attributed to Arthur Samuel's work in 1959 (29)). This review will not examine non-machine learning approaches to AI.

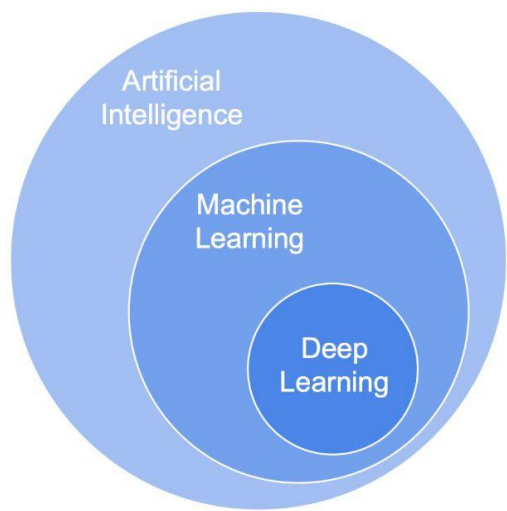

**Figure 1. Terminology.** *Deep learning is a subfield of Machine Learning, which is a subfield of Artificial Intelligence. This image was adapted from [Goodfellow et al.](Chapter 1, Page 9)*

Most machine learning algorithms can be classified as being supervised, unsupervised, or reinforcement learning approaches. In supervised learning, observations in the dataset need 'ground truth' labels, and the algorithm learns to identify patterns in the data that are indicative of a certain label. Unsupervised approaches learn useful properties of the dataset, such as clustering (e.g., deriving ways of phenotyping sepsis (30)), without requiring any labels. Reinforcement learning allows the use of trial and error to learn which actions generate the greatest rewards. Supervised learning algorithms tend to be used most frequently, and are the focus of this article.

In the supervised learning paradigm, each *training example*, or observation, in the dataset, has two key components: its *features* (typically termed 'variables' or 'predictors' by epidemiologists) and its *label* (typically termed 'outcomes' by epidemiologists). Based on internal parameters, the learning algorithm processes the features and produces an output, which can be compared to the ground truth label. As the learning algorithm views more training examples, it adjusts its parameters to minimize the difference between it's output and the ground-truth labels. These differences are quantified using a *loss function*, which is chosen based on the type of task the model is performing. The further the model's output from ground-truth, the greater the loss. A perfect model would have a loss of zero. This may sound quite familiar; linear regression and logistic regression are both supervised learning algorithms.

In addition to regression, numerous other machine learning algorithms fall outside the scope of deep learning, including support vector machines, naive-Bayes algorithms, and decision trees (Bi and colleagues provided a thorough review of these methods in an earlier volume of AJE (31)). Terms used in the field of machine learning tend to vary from those used in the medical literature. Table 1 contains common machine learning vocabulary and analogous epidemiologic terms as a reference for the reader.

**Table 1.** Analogous epidemiologic terms for common machine learning vocabulary

| Machine learning term | Analogous epidemiology term or concept |
|---|---|
| Training example | Observation, individual |
| Feature | Predictor, covariate, independent variable |
| Label | Outcome, response variable |
| Noisy labels | Outcome with measurement error |
| Feature engineering | Data pre-processing |
| Weights | Model coefficients |
| Bias term | Model intercept |
| Training | Model fitting |
| Training set | Derivation set |
| Bagging | Bootstrap model selection |
| Model output | Prediction |
| Sigmoid classifier | Logistic regression |
| Softmax classifier | Multinomial logistic regression |
| L1 regularization | LASSO regression |
| L2 regularization | Ridge regression |
| Confusion matrix | Contingency table, 2x2 table |
| Recall | Sensitivity |
| Precision | Positive predictive value |

# DEEP LEARNING

Deep learning is a collection of machine learning methods, where stacked processing layers are used to create abstract representations of data, creating an artificial "neural network" (32,33). These processing layers form an interconnected path from input features (e.g. age, smoking status, systolic blood pressure, etc.) to the model's output (e.g. risk of heart disease). Initially inspired by mechanistic theories of brain physiology, each layer consists of processing units called 'neurons' or 'nodes'.

Neural networks may be considered a more flexible approach to fitting a prediction model. Prediction modeling typically involves numerous pre-processing decisions: selecting a subset of the available predictors using prior knowledge or a statistical procedure (e.g. stepwise regression), discretizing continuous predictors, introducing polynomials, or using interaction terms. In deep learning, with enough data, pre-processing steps become unnecessary because of the flexibility gained by multiple processing layers and nonlinear data transformations, known as activation functions (32) (Figure 2).

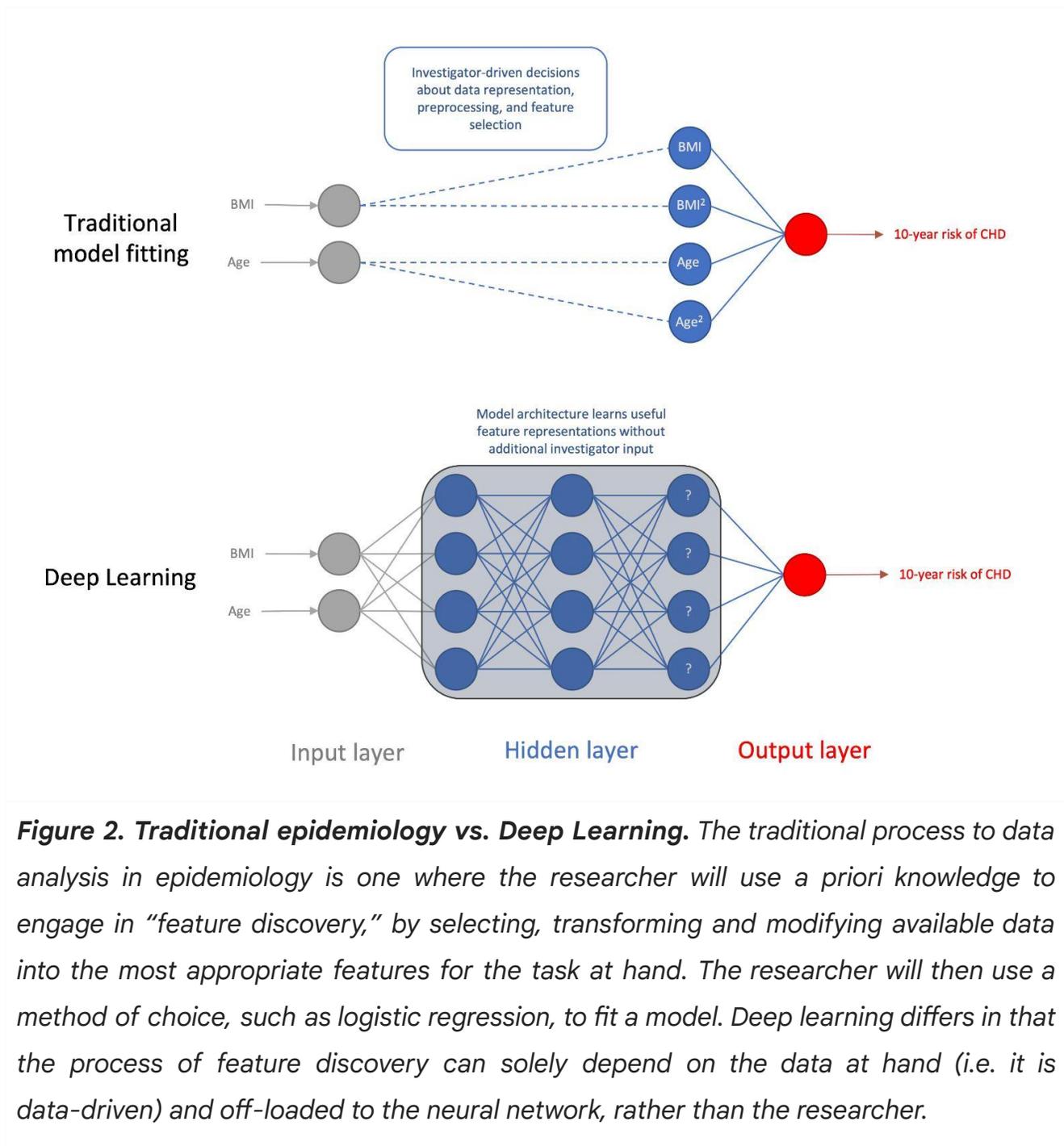

**Figure 2. Traditional epidemiology vs. Deep Learning.** *The traditional process to data analysis in epidemiology is one where the researcher will use a priori knowledge to engage in "feature discovery," by selecting, transforming and modifying available data into the most appropriate features for the task at hand. The researcher will then use a method of choice, such as logistic regression, to fit a model. Deep learning differs in that the process of feature discovery can solely depend on the data at hand (i.e. it is data-driven) and off-loaded to the neural network, rather than the researcher.*

*Structure*

The building block of neural networks are neurons, each of which perform two fairly simple operations. First, the neuron calculates a weighted sum of the inputs; these weights are randomly chosen at the start of model training and progressively revised to improve the performance of the model (i.e., minimize loss) throughout the learning process. During the second step, the neuron applies a non-linear mathematical transformation, an activation function, to that weighted sum.

While both of their operations are simple, neurons can be extremely powerful in aggregate. By adding more layers (and more neurons per layer), it is possible to model highly complex functions, including non-linearities and interactions, without any further specification (Figure 3). The universal approximation theorem (34–36) demonstrates that a sufficiently deep (i.e., many layers) or wide (i.e., many neurons) neural network can approximate any continuous mathematical function.

A regression model can be expressed as a simple neural network. A network with an input layer, an output layer, and no hidden layers (or, hidden layers of neurons applying linear activation functions) will behave equivalently to linear regression.

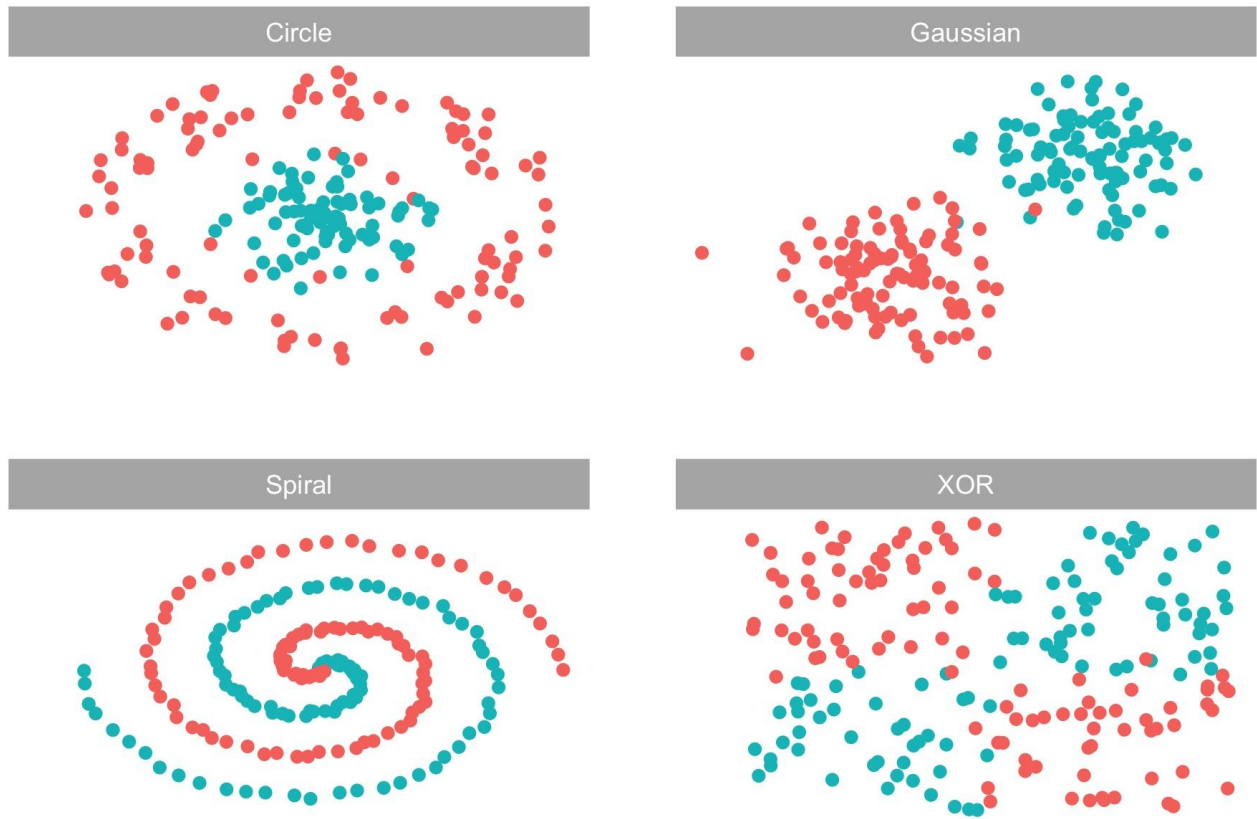

**Figure 3. Classification in increasingly non-linear data.** *In practice, discovering extremely complex high-dimensional functions is limited by finite amounts of data. Say we are interested in fitting a model to classify each dot into the red group or the green group based on its location. Now consider a simple regression model with one parameter per predictor. Such a model would only perform well in a setting where the underlying function is itself roughly affine; it lacks the capacity to model more complicated functions. We could increase its capacity by using splines, polynomials, or interaction terms, but even with such approaches, performance in say the spiral panel would be terrible. In deep learning, such problems can be addressed by adding more layers, adding more neurons per layer, changing the learning rate and other modeling choices collectively known as hyperparameters. You can test the impact of such choices in [Google Playground](#), from which the data were adapted.*

## Activation functions

After the weighted sum of the inputs is calculated by the neuron, an activation function applies a mathematical transformation. In theory, both linear and non-linear transformations may be used, but nonlinear functions are particularly useful because they increase the network's capacity to model nonlinearities in the data. In fact, it can be shown mathematically that multiple layers of linear transformations can be collapsed into a single layer.

The rectified linear unit (ReLU) is a simple activation function used extensively in the machine learning literature. If the input is positive, it outputs the value of the input. If the output is negative or zero, it outputs zero.

## Hyperparameters and training

'Hyperparameter' is a term used to describe any modifiable or "tunable" modeling choice. In regression, one could consider the use of higher order terms or interaction terms to be hyperparameters. In addition to increasing the complexity of the model structure, deep learning also increases the number of hyperparameters: the number of layers, the number of neurons in layers, parameters for regularization, and batch size (i.e. how many examples are shown to the model before updating weights), among others. These hyperparameters define the structure of the neural network and dictate how it will be trained. Values of hyperparameters can have a large impact on model performance, and should be reported to enhance reproducibility.

Among the most important hyperparameters is the learning rate. Unlike regression, loss functions cannot be formulaically minimized in deep learning models. Instead, an iterative approach known as gradient descent is commonly used. In simple terms, calculating the gradient tells us in which direction we should adjust parameter values. The magnitude of the change made is partly determined by the learning rate hyperparameter. The learning rate is critical to the success of the gradient descent algorithm; too large a learning rate can result in a failure to converge and too small will make the model train slowly and inefficiently.

# DEEP LEARNING ARCHITECTURES

## Fully-connected Neural Networks

Feed-forward neural networks (FNN) (also known as fully-connected neural networks, multi-layer perceptrons, or dense neural networks) are the most fundamental type of deep learning networks. They consist of one or more fully-connected layers (Figure 4). In a fully-connected layer, each neuron receives the output of all neurons from the previous layer. Based on learned weights, neurons calculate a weighted average of these inputs, apply an activation function, and propagate their output to neurons in the next layer.

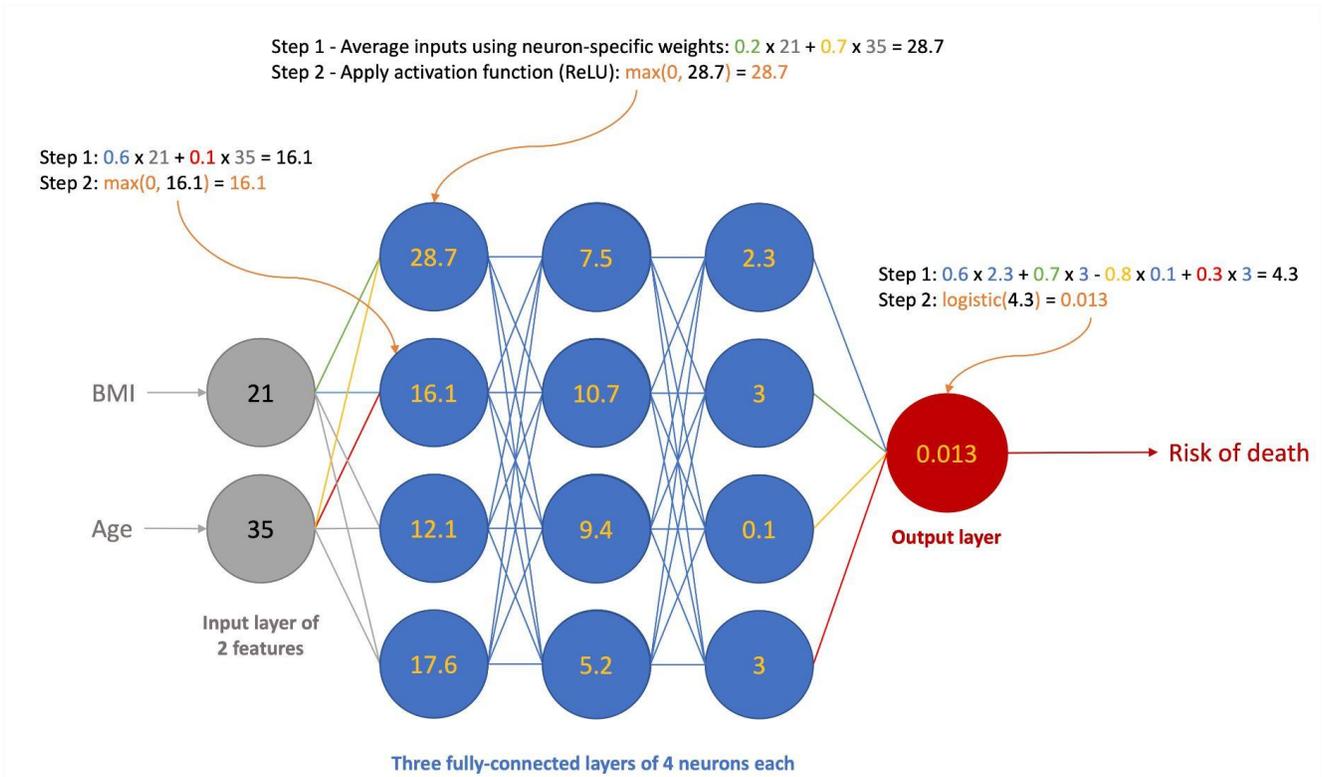

**Figure 4. The basic architecture of Fully-connected Neural Networks (FNNs).** *This figure demonstrates a FNN that predicts mortality on the basis of age (in years) and the body mass index (BMI; in kg/m2). FNNs only consist of fully-connected layers. This figure demonstrates three fully-connected layers, each of which consists of four neurons (in blue). Starting from the input layer (in gray), each neuron receives an input, takes a weighted average of that input using arrow-specific weights, applies an activation function (in orange; here we demonstrate ReLU: max(0, x)) and propagates the activated weighted average to each of the neurons of the next hidden layer. The depicted FNN is a 4-layer neural network because it consists of 4 layers of learned parameters (3 hidden layers, 1 output layer).*

## Examples in health research

Avati et al. (37) used the electronic health record (EHR) of 221,284 patients and 13,654 different features to predict all-cause mortality within the following year. To do so, they split the available data into training, validation and test sets at a ratio of 8:1:1. They trained an FNN with 18 fully-connected layers of 512 neurons each. The neurons in these layers used an activation function closely-related to the ReLU, a scaled exponential linear unit (SELU). The output layer used a single neuron with a logistic activation function to output a probability. Their model correctly identified 1 in 3 deaths at the pre-specified tolerance of 1 in 10 false alarms.

## Convolutional neural networks

Convolutional neural networks (CNNs) are a family of deep learning models particularly well-suited to tasks involving images. However, they can also be used with other data types (e.g., medical records) (38,39). The networks have three core components: convolutional layers, pooling layers and fully-connected layers. These layers can be rearranged into different architectures of varying complexity.

### *Convolutional layers*

Generally speaking, there are several reasons why FNNs are poorly suited to process images. They require an inefficiently large number of trainable parameters. Digital images are represented to machine learning models as grids of tens of thousands of pixels, each with a numerical red, green and blue (RGB) value. If each RGB value is a feature and each neuron is fully connected, the exponential increase in weights from adding neurons or layers quickly becomes an issue. Further inefficiencies arise because weights learned by neurons are not shared in FNNs; if the network learns to locate an object of interest in one area of the frame, it will need to re-learn the pattern if the object is shifted. Similarly, FNNs lack an inherent structure to compare a given pixel with the pixels around it. Finally, an FNN is only capable of processing images of a fixed size.

These issues motivated the creation of convolutional layers (Figure 5). Conceptually, convolutional layers recognize patterns across an image by maintaining a consistent but small number of weights. Each convolutional layer is a square (e.g., 1x1, 3x3, 5x5) of learned weights, known as the 'filter'. The filter is first applied at the top left of an image, and the weight of the filter corresponds to a pixel. Each convolutional layer can have multiple filters, the same way a fully-connected layer can have multiple neurons. A pixel's value is multiplied by the corresponding weight. The filter then moves to the right by a pre-specified number of pixels or 'stride,' and repeats the process of multiplying pixel values by weights to create an image representation. Continuing this process across the whole image is a 'convolution.' As with fully-connected layers, an activation function is applied to the output of the filter before being fed to the next layer.

Convolutions create 'translation invariant' representations, meaning their output is consistent regardless of where in the picture the object may be. Often, they are conceptualized as 'feature detectors' because they are able to represent specific features - early layers detect basic features, such as edges and outlines of a face. Later layers detect more complex features and shapes, such as the eyes or the mouth of a face.

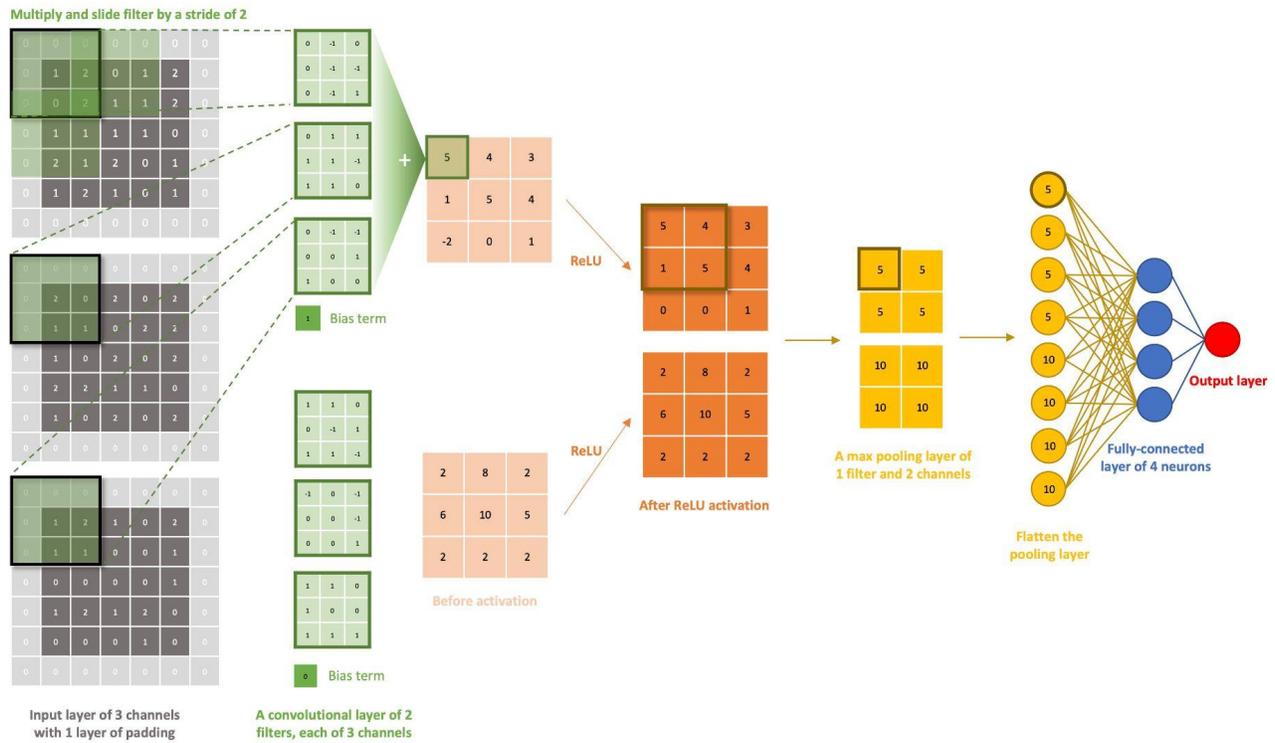

***Figure 5. The basic architecture of Convolutional Neural Networks (CNN).*** *CNNs are characterized by their inclusion of convolutional and pooling layers. Each of these layers consists of one or more filters of identical dimensions and of as many channels as their input. The input in this figure consists of three channels, in reminiscence of the standard RGB channels of an image (but, the input does not need to be an image and does not need to be constrained to 3 channels). Each channel of each filter is applied to the top left corner of the input, each input is multiplied by its respective weight and all values are then summed across channels, including the bias term. The filter is then moved by a predetermined stride (i.e. the number of squares by which the filter will move) to the right and down to cover the whole input. A padding of zeros is used so that the filter always perfectly fits into the input. An activation function is then applied to each of the totals, filter-specific outputs are stacked onto each other (such that the output of each filter now represents a new channel) and the stacked output is propagated to the next layer. A convolutional layer is almost always followed by a pooling layer (in this case and in most cases, max pooling), which is applied across the output of the convolutional layer in a similar fashion as described above. Note that no activation is applied after max*

*pooling. Finally, it is common to finish off a CNN with one or more fully-connected layers, where each rectangle of the max pooling output represents a single feature input to the fully-connected layer. Image adapted from the [CS231n course](#) at Stanford University.*

## Pooling layers

Convolutional layers are interlaced with pooling layers, which aggregate information across rectangular "neighborhoods" of an image. This reduces the size of its representation and helps neural networks identify specific features (e.g. a mouth of a face), regardless of their location within the image; they play a critical role in the performance of CNNs (40). Typically, pooling layers will take the maximum value of the neighborhood ("max pooling"), but they can also take the average ("average pooling"). Unlike most other types of layers discussed, no activation function is applied to the output of pooling layers, nor do they contain any learned parameters.

## Examples in health research

In 2017, Esteva et al. published a paper evaluating the ability of a CNN to classify skin lesion photographs into the risk of having each one of 2,032 different dermatologic diseases (41). Authors found the model achieved performance on par with 21 board-certified dermatologists in prediction of keratinocyte carcinomas versus benign seborrheic keratoses and malignant melanomas versus benign nevi.

The study used a previously-developed CNN architecture, Inception-v3 (42). The network contained repeated convolution and pooling layers, and had been originally trained to classify ImageNet, a non-medical dataset of over 14 million photographs from more than 21,000 categories (12,43). Esteva et al. leveraged this "pre-trained" model, and refitted the final fully-connected layers using the dermatology images and labels in their dataset. Repurposing pre-trained CNNs for medical tasks is a frequently-used strategy known as 'transfer learning'; otherwise, training performant CNN architectures from scratch can require enormous resources.

## Recurrent neural networks (RNNs)

There are many tasks where the ordering of model inputs matters, such as the sequence of words in natural language processing or the sequence of notes in music recognition. Recurrent Neural Networks (RNNs) process data in a sequential fashion, achieving better-than-human ability in tasks such as speech recognition (44).

### Recurrent layer

In its simplest form, a recurrent layer is a fully-connected layer that feeds into itself; outputs at one timestep become inputs at the next timestep (Figure 6). Take for example the phrase "the quick brown fox jumps over the lazy dog". We first create a mathematical representation for each word. A straightforward way of doing so is to create an indicator vector (i.e., a vector where the position corresponding to the specific word equals to 1 and all other positions to 0). However, the indicator method is inefficient; words can alternatively be represented as embeddings, where words that are similar are represented by vectors that are close to one another in vector space. Using embeddings tends to improve model performance, and a number of these embeddings are open sourced and freely available (10,45,46).

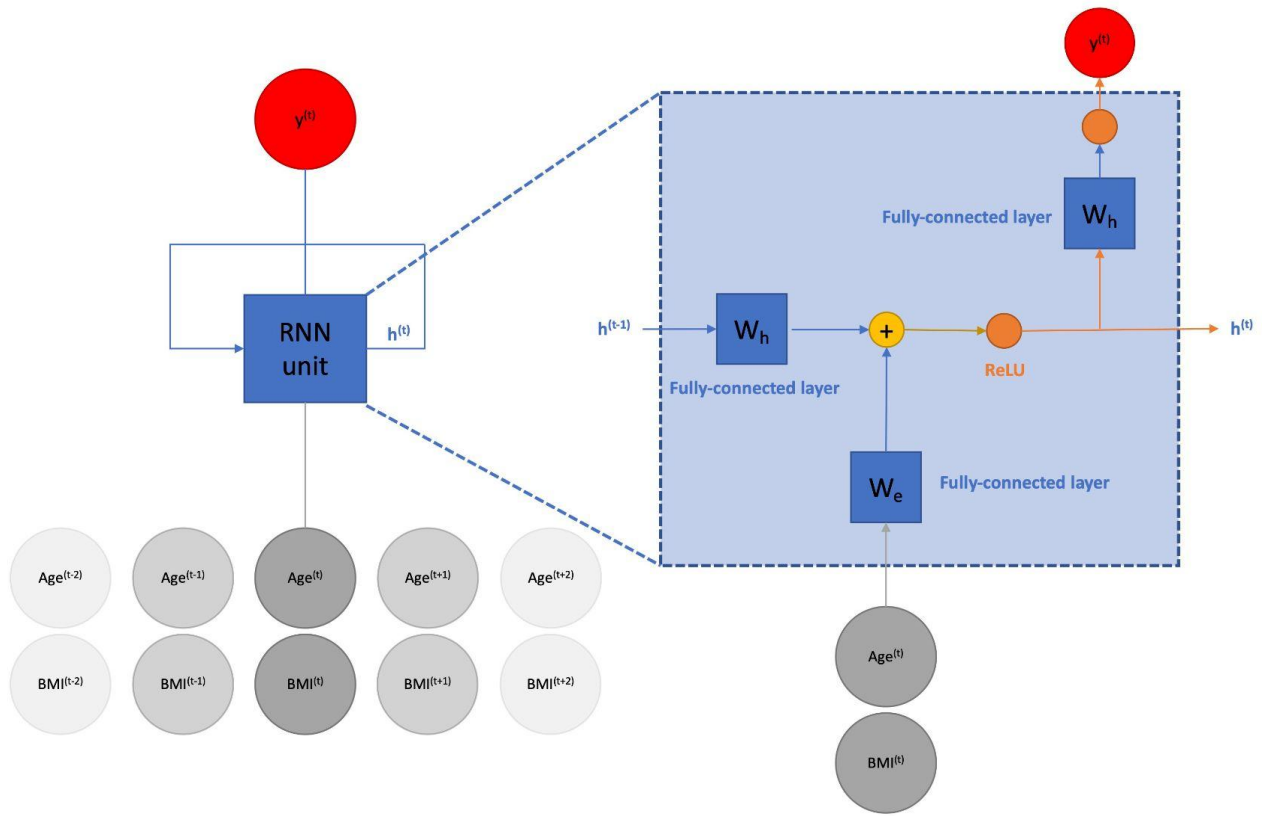

*Figure 6. The basic architecture of Recurrent Neural Networks (RNN).* RNNs are characterized by their inclusion of recurrent layers. Each recurrent layer takes a weighted average of input from the previous time point (known as the "hidden state"), adds it to a weighted average of the input from the current time point (known as the "embedding"), adds a bias term, applies an activation function (e.g. ReLU) and propagates the activated weighted average to the next time point. Image adapted from Thamarasee Jeewandara's *article (2020) on Phys.org.*

A basic RNN unit consists of a single layer and an activation function, and it will process the sequence from left to right. The vector representing the first word, 'the' is propagated through the layer, and an activation function is applied. The output will be concatenated (i.e. bundled) with the vectorized representation of "quick", passed through the same network, and the output will again be concatenated with the representation of the next word. This process continues as the model iterates through the entire sequence. Notice that the RNN unit remains unchanged, we are simply looping through the inputs. Predictions or inferences can be made using outputs of the model at any stage (i.e., a prediction can be made after each item in the sequence or after the entire sequence has been processed).

Simple RNN units often fail to retain key contextual information that occurred earlier in the sequence (47). Instead, many RNNs use other types of units, such as Gated Recurrent Units (GRU) (48), or Long-Short Term Memory (LSTM) units (49). LSTMs have an additional 'memory' state that allows storage of information from previous hidden states, in addition to the hidden state maintained by the simple RNN unit. At each subsequent time step along the sequence, the LSTM takes two inputs: the hidden state from the previous time step and the values of the predictors at the current time step. Using these inputs, it determines what new information to retain, what old information to forget and what information to pass along to the next time step as an output. This last value is the new hidden state at the current time point and captures information that was seen recently, as well as several time steps before.

### *Examples in health research*

Use of RNN architectures has often led to higher performance than standard feedforward networks (14). Choi et al. used an RNN to predict the risk of being diagnosed with heart failure in a matched case-control sample within the next 12 months (50). Embeddings were used to capture clinical events in the timeline of a patient (e.g. being diagnosed with pneumonia, having a chest X-Ray, or being prescribed amoxicillin). A GRU was used to propagate through each event in the sequence it was recorded. At the end of the sequence of clinical events, the model output a probability of heart failure.

*Modern directions*

There has been substantial work on alternative architectures for sequences that mitigate some of the limitations of RNNs. RNNs cannot organically represent events occurring at varying or irregular time intervals; recent work proposed RNNs that work on a continuum, rather than discrete event units (51). Even LSTMs and GRUs have limited capacity to effectively capture long-term dependencies; transformer models use a mechanism called 'attention' (52) to simultaneously process all sequence elements, which has led to enormous improvements in performance (8,10).

## Alternative architectures

CNNs and RNNs represent only a portion of deep learning architectures. Other models include deep generative models (53), Bayesian neural networks (54,55), graphical models (56), and general adversarial networks (57–59), though we are unable to describe these architectures in detail in this review.

## MODEL FITTING

Deep learning models can have substantial capacity. Given raw data and labels, the network can "discover" intermediate representations that facilitate translation of a given input into the desired output, referred to as 'end-to-end' learning. Neural networks have the ability to use, combine, and create intermediate representations of features through learned weights. Figuring out which models are 'optimal' typically requires partitioning the data, fitting multiple models with different specifications, and choosing the best-performing model according to a metric of interest.

## Training, validation, and test sets

Using the same data to fit a model and evaluate it will lead to an overoptimistic estimate of performance in the target population. To prevent this, machine learning typically splits the available data into three sets: the training, validation, and test sets (the training-validation split can be avoided if cross-validation is used). The split ratio can vary depending on total dataset size. The training set can be used to train a variety of models. Performance of these models is compared by measuring their performance on the validation set. The 'held-out' test set should be used *only once*, to assess performance after the final model has been selected.

## Hyperparameter tuning

There is typically limited theoretical understanding of which modeling choices will work well for a given machine learning task. Choosing hyperparameters often requires iterative experimentation, a process known as tuning. Candidate values can be found through random search (values randomly selected independently of one another) or Bayesian hyperparameter optimization (60) (conditioning on the observed performance of previous hyperparameter combinations to inform which combination of hyperparameters should be tested next).

## Regularization

The ultimate goal of fitting a model is to generalize well to a target population that has not yet been "seen" by the model. Models with sufficient capacity can 'memorize' training data - they fit to noise rather than signal - and become 'overfit.' In contrast, models with inadequate capacity will poorly model the underlying function; this is often referred to as 'underfitting'. Overfitting is diagnosed when a model performs very well in the training set, but poorly in the validation set. Underfit models will perform poorly both in the training and validation sets.

Regularization helps to minimize overfitting. There are several methods to regularize models; the most widely used is weight decay, which adds an extra term to the loss function to shrink small weights to 0 (known as L1 regularization) or to shrink all weights by a small amount (known as L2 regularization). In addition to these methods, deep learning utilizes many other methods of regularization, such as adding random noise (61) to the inputs, dropout (62) and batch normalization (63)

## Performance metrics

A number of metrics can be used to quantify the performance of the model and appropriate performance metrics will depend on the task. For binary classification tasks, common metrics include recall (sensitivity in epidemiological literature), precision (positive predictive value in epidemiological literature), accuracy, area under the receiver operating curve (AUROC), area under the precision-recall curve (AUPRC) and calibration.

AUPRC is a particularly useful metric for measuring performance when there is a large imbalance in the prevalence of different outcome labels (64,65). It is a measure of average positive predictive value, across all values of sensitivity and varies in value from the prevalence of the outcome in the sample (no predictive ability) to 1 (perfect predictive ability).

Calibration measures how well the expected risk corresponds to the observed risk and may be assessed using a calibration curve, the Greenwood-Nam-D'agostino test (a modified version of the Hosmer-Lemeshow chi-squared statistic) (66) or the Brier score.

Reporting multiple metrics with confidence intervals and a clinically meaningful interpretation of the result is generally good practice. On their own, standard machine learning metrics do not quantify potential benefits or harms to the patient.

*Practical challenges for real-world use of deep learning*

The use of machine learning-enabled technologies in real-world clinical settings presents numerous practical challenges (67).

In non-healthcare settings, model performance tends to degrade with time, and the performance originally measured in the test set overestimates real-world performance. This phenomenon is known as the "training-serving skew." We can view this as a failure of the model to generalize to the population it is currently being used in (a lack of external validity). This can have several root causes, including temporal shifts in the input features (e.g., changes in patient behavior, clinical or operational practices), or differences between the population included in the training set versus actual users of the technology. Continuous performance monitoring is important for detection of performance degradation, and deployed deep learning models are often retrained on fairly regular schedules to mitigate this issue.

The use of machine learning in clinical settings also has the potential to propagate or increase existing disparities in healthcare. Ensuring fairness requires a holistic approach (68), including consideration of biases in formulation of the machine learning task, training set composition, labeling of observations, non-random missingness of data, and real-world use of the models.

## CONCLUSION

In a 1970 commentary, physician William Shwartz mused, "indeed, it seems probable that in the not too distant future the physician and the computer will engage in frequent dialogue, the computer continuously taking note of history, physical findings, laboratory data, and the like, alerting the physician to the most probable diagnoses and suggesting the appropriate, safest course of action." (69) While this vision has certainly not been realized, deep learning may enable tooling that facilitates parts of it. Deep learning presents real opportunities for improving the quality of care provided to patients, as well as numerous challenges.

We hope this review has provided you with the foundation needed to engage with research that uses deep learning, either as a collaborator, reviewer, or critical reader. Epidemiologists have a critical role to play in the development of these technologies and in measuring their real-world impact and safety.

# Acknowledgements

## Author affiliations

1. Google LLC, Mountain View, California (Stylianos Serghiou)

2. Meta-Research Innovation Center at Stanford, Stanford University School of Medicine, Stanford University, Stanford, California (Stylianos Serghiou)

3. IQVIA (Kathryn Rough)

## Author contribution

Both authors contributed equally to this work.

## Financial support

This work was supported by Google, LLC.

## Presentation at a meeting

A version of this work was presented at the Society for Epidemiologic Research's 2019 Meeting.

## Conflicts of Interest

Stylianos Serghiou and Kathryn Rough were employed by Google, LLC at the time this article was originally drafted. Kathryn Rough currently is employed by IQVIA.